\documentclass{article}

\usepackage{PRIMEarxiv}

\usepackage[utf8]{inputenc} % allow utf-8 input
\usepackage[T1]{fontenc}    % use 8-bit T1 fonts
\usepackage{hyperref}       % hyperlinks
\usepackage{url}            % simple URL typesetting
\usepackage{booktabs}       % professional-quality tables
\usepackage{amsfonts}       % blackboard math symbols
\usepackage{nicefrac}       % compact symbols for 1/2, etc.
\usepackage{microtype}      % microtypography
\usepackage{lipsum}
\usepackage{graphicx}
\graphicspath{{media/}}     % organize your images and other figures under media/ folder

\title{Better Question-Answering Models on a Budget
}

\author{
  Yudhanjaya Wijeratne \\
  Backyard Labs \\
  Colombo, Sri Lanka\\
  \texttt{yudhanjaya@appendix.tech}\\
   \And
  Ishan Marikar\\
  Backyard Labs \\
  Colombo, Sri Lanka \\
  \texttt{ishan@appendix.tech} \\
}

\begin{document}
\maketitle

\begin{abstract}
Low-rank adaptation (LoRA) and question-answer datasets from large language models have made it much easier for much smaller models to be finetuned to the point where they display sophisticated conversational abilities. In this paper, we present Eluwa, a family of LoRA models that use the Stanford Alpaca dataset and massively improve the capabilities of Facebook's OPT 1.3B, 2.7B and 6.7B models. We benchmark these models in multiple ways, including letting GPT-4 judge their answers to prompts that span general knowledge, writing, programming and other tasks. We show that smaller models here can be fine-tuned to be as performant as models 3x larger  - all for as little as 40 USD in compute.
\end{abstract}

\section{Introduction: A brief history of recent and fortunate events} %% {{{

This section will serve as a very brief summary of recent milestones in the space. To wit:

Fine-tuning large language models has always been an expensive task, with the current state of the art requiring investments in the millions of dollars. 

In 2021, Low-Rank Adaptation (LoRA) \cite{hu2021lora} was introduced. It was a way of freezing a pre-trained model and injecting smaller weights into it, drastically reducing the number of parameters one had to train and the memory required to do this in. It made it possible to fine-tune enormous, expensive language models (like GPT-3) without needing the hardware or budget used to make those models in the first place. \\

In late 2022, the Self-Instruct paper \cite{wang2022self} showed that you could get a language model to be better at instruction-following ability by bootstrapping it off its own responses. It was an elegant way of creating datasets that could be used for fine-tuning models into impressive improvements. \\

In 2022, Meta AI rolled out OPT, a series of models that attempted to replicate GPT-3 \cite{zhang2022opt}. These were expensive models to train (by our standards: OPT needed 992 80GB Nvidia A100 GPUs). Early in 2023, they released LLaMA \cite{touvron2023llama}, a collection of language models ranging from 7B to 65B parameters. The 13B model was particularly interesting, as it was demonstrated to outperform OpenAI's much-discussed GPT-3 175B model (the 'B' here all stand for \textit{billion}, as in \textit{billions of parameters}). By this point Meta had gone from replication to making their open projects vastly more efficient.\\

\newpage
Which is where the Stanford Alpaca project\cite{alpaca} enters the fray. Using a dataset of 52,000 question-answer pairs, generated in the manner demo'd in the Self-Instruct paper, these researchers managed to take the LLaMA 7B model to GPT3's text-davinci-003 levels. It neatly coincided with a wave of improvements in the field of 8 and 4-bit quantization by Dettmers et al (see work from 2015\cite{dettmers20158} to now \cite{dettmers2022case}), which not only made it possible to fit models into smaller amounts of RAM and VRAM, but also put the means to do so directly into the popular Huggingface Transformers library.

This was a monumental achievement, because it essentially put GPT-3 capabilities into the range of consumer hardware; see Gerganov's llama.cpp\cite{llamacpp} for how far this type of technology can go.\\

This accessibility set off a veritable Cambrian explosion of camelid-themed model improvements. \\

There is, for instance, Cabrita \cite{cabrita2023}, a Portuguese Alpaca trained with a translation of the Alpaca data. Likewise there is Japanese-Alpaca \cite{japanesealpaca}. \footnote{On a semi-related note, our attempt at creating a Sinhala Alpaca using the same methods failed.}

There is Vicuna, an even further improvement on LLaMA \cite{vicuna2023}, which rather interestingly was rated using OpenAI's GPT-4. There is GPT4all\cite{anandgpt4all}, which someone has even gotten up and running on a TI-84 calculator\footnote{See https://www.youtube.com/watch?v=9WXIAsbTMZ0. By the time I finish this paper there may be several other animals being released on Github, and at least one of them might run on someone's cellphone.}. 

Which brings us to our experiment here. Much work has been done on Llama. But can other models be similarly improved - especially ones that are smaller and better understood? 

%% }}} --- Section

\section{Methodology} %% {{{

We take a little bit of everything spoken about for this project. We loaded up a series of OPT models (1.3B, 2.3B and 6.7B). To keep costs low (in compute requirements, and thus in actual dollars), we used Low-Rank Adaptation and trained them on the Alpaca dataset. \\ 

For each OPT model, we trained two LoRA models - one at 1000 iterations with our training parameters, the other at 2 full epochs. Our training parameters were as follows: \\
\begin{verbatim}
per_device_train_batch_size=8
gradient_accumulation_steps=4
warmup_steps=100
learning_rate=2e-4
fp16=True
logging_steps=10
\end{verbatim}

Varying the learning rate from 1e-4 to 3e-4 did not make much of a difference to the loss curve; ultimately we settled on the figure used by the Alpaca project.

We deliberately stuck to low-resource parameters: we did the training on Google Colab, and all these models fit within a GTX 1080 Ti. \\ 

The final product is the Eluwa family of models. Eluwa means 'goat' in Sinhala \footnote{As pointed out, there is now a mild trend of naming LLaMA-derived models after other camelids.
 Well, in Sri Lanka, we don't have llamas (at least, I've never seen any), but we do have goats. "Eluwa" means goat. Goats are fearsome, versatile, and double as the essential ingredient in mutton rolls. Everything in the known universe is either a goat, or not a goat. They're not as nice as llamas or alpacas, but they'll do.}.
%% }}} --- Section

\newpage

%% ---------------------------
%% A section
%% ---------------------------
\section{Testing Eluwa} %% {{{

We went about testing Eluwa two ways. First, we adopted the Vicuna benchmark: 80 questions under various categories, ranging from questions like \textit{"Can you explain the basics of quantum computing?"} to \textit{How do cultural, social, and economic factors influence people's food choices, and how can this knowledge be used to promote healthier diets?} to \textit{If you were a Shakespearean character, how would you declare your love for someone in a soliloquy?} to counterfactual questions like \textit{What if the Aztecs had successfully repelled the Spanish conquistadors?}. The answers to these questions were then scored by GPT-4, arguably one of the most sophisticated models publicly available as of the time of writing\footnote{This is by no means a comprehensive suite, but it's been used before to compare Vicuna to ChatGPT (GPT 3.5) across a broad range of tasks and questions of the kind that a chat model might receive.}.\\

Second, we tested Eluwa on Wikitext-2 (a small pre-processed sample of the Wikitext dataset\cite{merity2016pointer}. \footnote{While there are plenty of other benchmarks that are easily accessible, it turns out that HuggingFace's evaluation library hasn't quite yet caught up with 8-bit models running LoRAs. We were fortunate to find recent work by this project to clean up the Alpaca dataset: https://github.com/gururise/AlpacaDataCleaned/blob/main/README.md; that made it easy for us to test Eluwa on Squad and others.}. 

\subsection{The Vicuna Benchmark}

We used the following settings for all models this test: 

\begin{verbatim}
top_p=0.70
top_k=0
temperature=1.0
repetition_penalty=1.1
typical_p=1.0
num_beams=1
\end{verbatim}

Inference was performed on a single GTX 1080Ti. GPT-4 was prompted with the following phrase:

\textit{"You are a sophisticated and critical judge evaluating the outputs 
of AI models. For each question given above, evaluate the "answer" 
as a response to the prompt. Judge each answer on a score of 1-10
based on relevance, coherence, and on its informational value as 
a response to the question. Actions should only be considered useful 
if their output is also present in the answer. Generate both this 
score and your reasoning for the score.  Format your response as 
a table containing question ID, score, and reasoning."}\footnote{

We had to include the '\textit{Actions should only be considered useful if their output is also present in the answer}' clause because the OPT family of models seem to have a tendency to say some variant of "Sure! I'll start work on it right away." in response to prompts that they can't quite deliver on.} Under this schema, we obtained the following results:

\begin{table}[!ht]
    \centering
    \begin{tabular}{|l|l|l|l|}
    \hline
        Model & OPT 1.3b base & Eluwa 1.3b 1000 iter & Eluwa 1.3b 2 epoch \\ \hline
        Generic & 53 & 58 & 59 \\ \hline
        Knowledge & 71 & 69 & 78 \\ \hline
        Roleplay & 32 & 44 & 54 \\ \hline
        Common sense & 41 & 55 & 57 \\ \hline
        Fermi & 21 & 19 & 17 \\ \hline
        Counterfactual & 28 & 29 & 42 \\ \hline
        Coding & 16 & 7 & 8 \\ \hline
        Math & 3 & 3 & 3 \\ \hline
        Writing & 22 & 11 & 54 \\ \hline
        Total & 287 & 295 & 372 \\ \hline
    \end{tabular}
\end{table}

\begin{table}[!ht]
    \centering
    \begin{tabular}{|l|l|l|l|}
    \hline
        Model & OPT 2.7b base & Eluwa 2.7b 1000 iter & Eluwa 2.7b 2 epoch \\ \hline
        Generic & 22 & 44 & 57 \\ \hline
        Knowledge & 35 & 60 & 72 \\ \hline
        Roleplay & 29 & 38 & 58 \\ \hline
        Common sense & 20 & 48 & 50 \\ \hline
        Fermi & 4 & 28 & 23 \\ \hline
        Counterfactual & 5 & 24 & 23 \\ \hline
        Coding & 2 & 7 & 7 \\ \hline
        Math & 0 & 3 & 3 \\ \hline
        Writing & 8 & 19 & 19 \\ \hline
        Total & 125 & 271 & 312 \\ \hline
    \end{tabular}
\end{table}

\newpage

\begin{table}[!ht]
    \centering
    \begin{tabular}{|l|l|l|l|}
    \hline
        Model & OPT 6.7b base & Eluwa 6.7b 1000 iter & Eluwa 6.7b 2 epoch \\ \hline
        Generic & 65 & 74 & 77 \\ \hline
        Knowledge & 57 & 89 & 72 \\ \hline
        Roleplay & 49 & 70 & 80 \\ \hline
        Common sense & 67 & 80 & 86 \\ \hline
        Fermi & 27 & 16 & 26 \\ \hline
        Counterfactual & 30 & 43 & 71 \\ \hline
        Coding & 9 & 9 & 7 \\ \hline
        Math & 3 & 3 & 10 \\ \hline
        Writing & 13 & 24 & 10 \\ \hline
        Total & 320 & 408 & 439 \\ \hline
    \end{tabular}
\end{table}

The Vicuna benchmark shows consistent improvement over the base OPT models. The 1.3b OPT model seems to be better, on the surface than the 2.7b model. We don't know why this is. 2.7b was more curt, more prone to insulting the user, refusing to elaborate on answers. It's interesting that the Eluwa 1.3 2-epoch model is doing better here than base OPT 6.7b (a model 5 times its size). However, a word of caution: 1.3's verbosity seems to come from regurgitation, and less refined ability to write, and indeed in our testing we found this family of models answering questions with fabricated blogposts, right down to authors' bios and twitter credentials. \\

\subsection{ Wikitext-2 results}

With the best models from the previous test selected, we then tested on Wikitext-2. We used the following settings for all models this test: 

\begin{verbatim}
temperature=0.1,
top_p=0.8,
top_k=40,
num_beams=1,
repetition_penalty=1.2
\end{verbatim}

Inference was performed on a Google Colab notebook. \\

\begin{table}[!ht]
    \centering
    \begin{tabular}{|l|l|}
    \hline
        Model & Wikitext Perplexity \\ \hline
        Eluwa 1.3b 2epoch & 14.296875 \\ \hline
        Eluwa 2.7b 2epoch & 12.0859375 \\ \hline
        Eluwa 6.7b 2epoch & 10.6015625 \\ \hline
        OPT 6.7b base & 10.28125 \\ \hline
        OPT 2.7 base & 11.78125 \\ \hline
        OPT 1.3 base & 13.8828125 \\ \hline
    \end{tabular}
\end{table}

This benchmark shows a slight increase in perplexity after training, indicating that the new models have a slightly more difficult time predicting the tokens of Wikitext-2. 

%% }}} --- Section
\section{Results}
The Eluwas, even with minimal work, were able to dramatically improve model outputs, such that Eluwa 2.7B is better here than the base OPT 6.7b model (a model almost 3 times the size). The 1.3b line is almost anomalously good, bordering on base OPT 6.7B. Thus we see a comfortable 30-50\% improvement in question-answering behavious within the same line of models; improvements that sometimes let models respond like models 3-5x bigger. \\

Secondly, there was nearly always an improvement in creative tasks - roleplaying (\textit{You are a mountain climber reaching the summit of Mount Everest. Describe your emotions and the view from the top.}), counterfactual thinking (\textit{What if the Suez Canal had never been constructed?}) and presentation-of-knowledge tasks. Given the dramatic improvements here, these models are perhaps undertrained for these types of tasks. \\

Coding, mathematics and vague writing tasks (\textit{Can you help me write a formal email to a potential business partner proposing a joint venture?}) did not improve as much. Nor did codewriting (\textit{Implement a Python function to find the longest common subsequence of two input strings using dynamic programming.}), or Fermi tasks, aka reasoning (\textit{How many snowflakes fall during a typical winter? Try to explain your answer. Your explanation should take the reader through your reasoning step-by-step.}). All models hallucinated heavily on these tasks. This is as expected.

This training and testing was completed for a total of about 40 USD (less in reality, as we wasted some money every time Google Colab disconnected on us). The final, most performant LoRA models are available for public use: all three models and the training code are available on Huggingface\footnote{https://huggingface.co/BackyardLabs} and on Github\footnote{https://github.com/yudhanjaya/Eluwa}.

%Bibliography

\bibliographystyle{unsrt}  
\bibliography{paper}

\end{document}